\documentclass[10pt,draftcls,journal,compsoc,letterpaper,oneside,onecolumn]{IEEEtran}
\usepackage[dvipdfm]{graphicx}  
\usepackage{cite}
\usepackage{url}
\usepackage[fleqn,cmex10]{amsmath}
\usepackage[psamsfonts]{amssymb}
\usepackage{listings}

\def\a{\boldsymbol{a}}

\def\vv{\boldsymbol{vv}}

\def\x{\boldsymbol{x}}
\def\y{\boldsymbol{y}}

\def\A{\boldsymbol{A}}

\def\D{\boldsymbol{D}}

\def\0{\boldsymbol{0}}

\def\wh{\widehat}

\begin{document}
\allowdisplaybreaks  

\title{%
  Fully automatic extraction of salient objects from videos in near real-time
}
\author{%
  Kazuma Akamine, Ken Fukuchi, Akisato Kimura and Shigeru Takagi
  \IEEEcompsocitemizethanks{%
    \IEEEcompsocthanksitem
    K. Akamine is with Department of Computer Science and System Engineering,
    Faculty of Engineering, Miyazaki University\hskip1em
    1-1 Gakuen Kibanadai-Nishi, Miyazaki, 889--2192 Japan.
    \IEEEcompsocthanksitem
    K. Fukuchi and S. Takagi are with Department of Information and
    Communication Systems Engineering, Okinawa National College of
    Technology\hskip1em Henoko 905, Nago, Okinawa, 905--2171 Japan.
    \IEEEcompsocthanksitem
    A. Kimura is with NTT Communication Science Laboratories, NTT
    Corporation,\hskip1em Morinosato Wakamiya 3-1, Atsugi, Kanagawa, 243--0198
    Japan. E-mail: akisato $<$at$>$ ieee org
    URL: http://www.brl.ntt.co.jp/people/akisato/
  }
}

\maketitle

\begin{abstract}
Automatic video segmentation plays an important role in a wide range of
computer vision and image processing applications. Recently, various
methods have been proposed for this purpose. The problem is that most of these
methods are far from real-time processing even for low-resolution videos due to
the complex procedures. To this end, we propose a new and quite fast method for
automatic video segmentation with the help of 1) efficient optimization of
Markov random fields with polynomial time of number of pixels by introducing
graph cuts, 2) automatic, computationally efficient but stable derivation of
segmentation priors using visual saliency and sequential update mechanism, and
3) an implementation strategy in the principle of stream processing with
graphics processor units (GPUs). Test results indicates that our method
extracts appropriate regions from videos as precisely as and much faster than
previous semi-automatic methods even though any supervisions have not been
incorporated.
\end{abstract}

\section{Introduction}
\label{sec:intro}

Extracting important (or meaningful) regions from videos is not only a
challenging problem in computer vision research but also a crucial task in many
applications including object recognition, video classification, annotation and
retrieval. It can be formulated as a problem of binary segmentation, where
important regions are considered ``objects'' and the remaining regions
``backgrounds''. One of the most promising ways to achieve precise segmentation
is the method proposed by Boykov et al. \cite{graphCuts:boykov1,graphCuts:%
boykov2} called Interactive Graph Cuts. This method originated in the work of
Greig et al. \cite{graphCuts:greig}, where the exact maximum a posteriori (MAP)
solution of a two label pairwise Markov random field (MRF) can be obtained by
finding the minimum cut on the equivalent graph of the MRF. Later, various
kinds of modifications, improvements and extensions have been presented in the
literature \cite{multiScaleGraphCuts:nagahashi,saliencyCuts:fu,conf_icme_%
FukudaTA08}. More recently, several approaches for extending it to video
segmentation have been proposed \cite{spatioTemporalGraphCuts:nagahashi,graph%
Cuts:kohliPaper}. In particular, Kohli and Torr \cite{graphCuts:kohliPaper}
described an efficient algorithm for computing MAP estimates for dynamically
changing MRF models, and tested it on the video segmentation problem.

\begin{figure}[t]
  \begin{center}
    \includegraphics[clip,width=0.7\hsize]{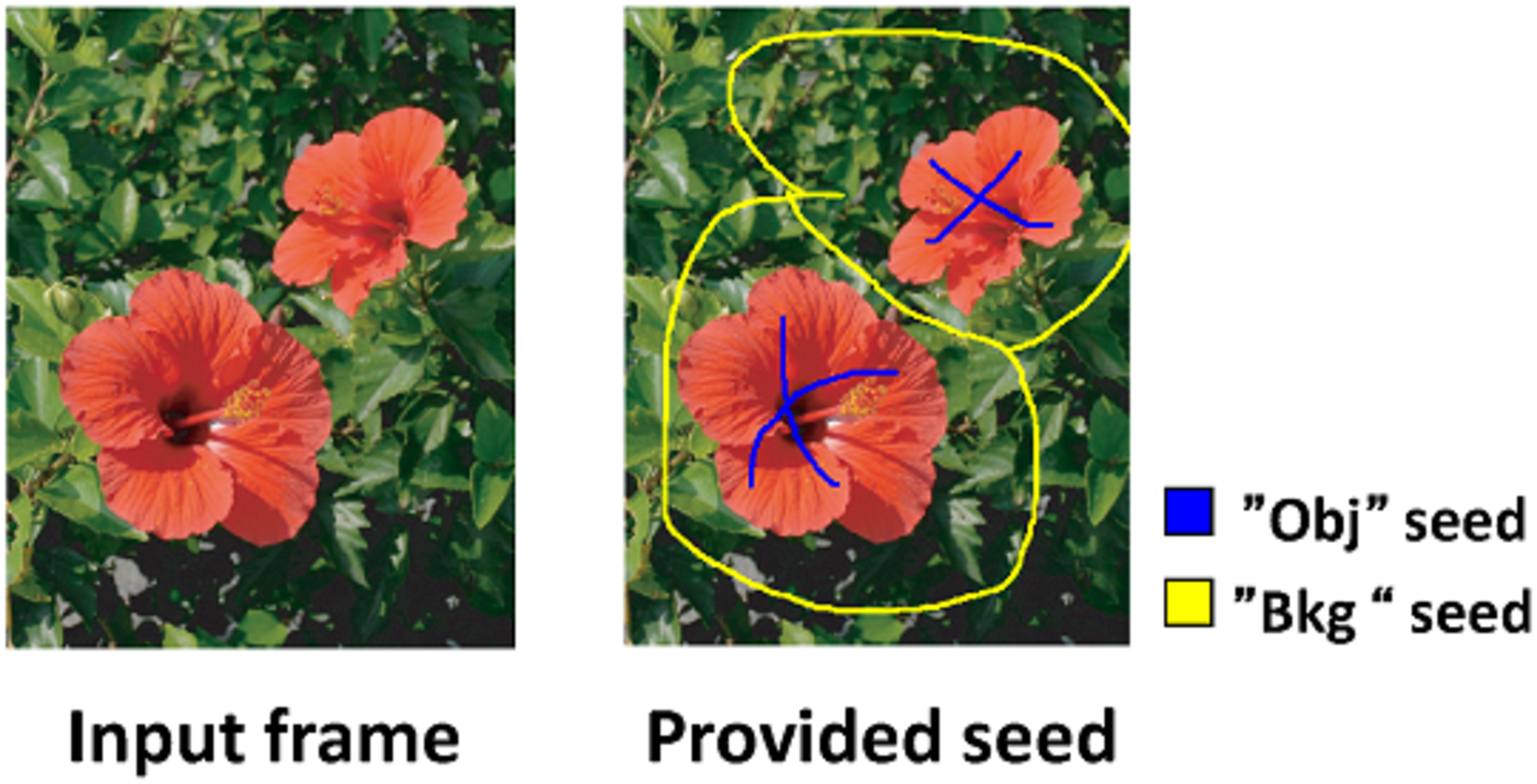}
    \caption{Example of manually-provided seeds}
    \label{fig:seed}
  \end{center}
\end{figure}

\begin{figure*}[t]
  \begin{center}
    \includegraphics[clip,width=\hsize]{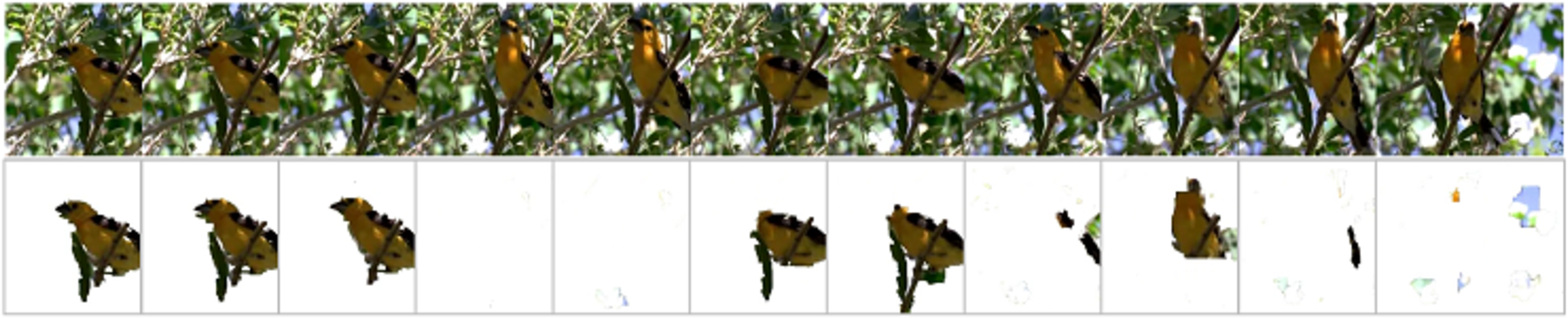}
    \caption{
      Example of a segmentation result (bottom) for some video (top) only with
      frame-wise saliency-based priors.
      We can see that segmented regions are frequently moved due to unstable
      behavior of visual saliency.
    }
    \label{fig:saliencyNotEnough}
  \end{center}
\end{figure*}

Although the above approaches are promising, they all pose a critical problem
in that they have to provide segmentation cues (seeds) manually and carefully
(See Figure \ref{fig:seed}). Such manual labeling is occasionally infeasible.
The development of fully automatic segmentation methods has been strongly
expected. Some previous work \cite{movingObjectDetection:Liu} utilized
motion information to achieve fully automatic detection and segmentation of
moving objects. However, targets we want to extract are not necessarily moving
in video frames; target objects might be traffic signs, nameboards or statues
that are all static. Also, a target object seems to be unmoving even though
it is actually moving since a video camera can appropriately pursuit the
target. Therefore, we need more versatile cues to extract various kinds of
targets.

The use of saliency-based human visual attention models is one of the
most promising approaches in this respect. The first biologically plausible
model for explaining the human attention system was proposed by Koch and Ullman
\cite{visualAttention:KochUllman}, and late implemented by Itti et al. \cite{%
Saliency:itti}. This model analyzes still images to produce primary visual
features (including intensity, color opponents, edge orientatiosn and
motion information), which are combined to form a saliency map that represents
the relevance of visual attention. Later, so many attempts have been made to
improve the Koch-Ullman model \cite{VisualAttention:Privitera,GeneratingEye%
Fixations:gu,VOCUS,saliency:minho2,decisionTheorySaliency:gao} and to extend it
to video signals \cite{decisionTheorySaliency:gao,DetectSurprise:itti,video%
Saliency:clement,saliency:minho1}. Our research group also proposed stochastic
models \cite{stochasticSaliency:derek,stochasticSaliencyMRF:derek} for
estimating human visual attention that tackled the fundamental problem of the
previous attention models related to the non-deterministic properties of the
human visual system. Such models would be helpful for automatically providing
segmentation seeds.

To this end, we propose a novel approach for achieving video segmentation based
on visual saliency. Our main contributions are as follows:

1) We newly incorporate saliency-based priors into frame-wise segmentation with
graph cuts to achieve fully automatic segmentation. For the purpose of still
image segmentation, this approach has been already appeared in the work
undertaken by Fu et al. \cite{saliencyCuts:fu}. However, when dealing with
video signals, segmentation results might be unstable (e.g. flickering or
frequent moving) due to fluctuations of visual saliency. Figure \ref{fig:%
saliencyNotEnough} depicts a segmentation result with saliency-based priors
derived from an input video. We can see from this figure that segmented regions
are frequently moved due to the instability of visual saliency. We have to note
that human visual attention might not be determined by only visual saliency
representing a kind of novelty calcuated only from image signals; human visual
attention is often controlled by their knowledge, experiences and intention.
This is the reason why there is a discrepacy between highliy salient regions
and intuitively attentive regions.

2) To tackle this problem, we develop a new technique for updating priors and
feature likelihoods, which makes use of another property of the human
visual system: temporal dependency of visual attention. We humans do not switch
our attention to various regions so frequently, even though salient regions
frequently move within a short period. Based on the above property, the new
technique additionally introduces the segmentation result obtained from the
previous frame to estimate a prior of the current frame. An idea of Kalman
filter is utilized to integrate the previous segmentation result and saliency-%
based priors and to obtain the actual prior density for segmentation.
Feature likelihoods can be also updated so as to reflect dominant feature
components of the previous segmentation result. Nevertheless the above
efforts, there still remains a crucial problem that it is still far from real-%
time processing due to its complex and costful procedures, especially in
estimating saliency-based visual attention, calculating feature likelihoods,
and deriving the segmentation results with graph cuts.

3) Thus, we introduce an implementation strategy making extensive use of stream
processing with graphics processor units (GPUs) to accelerate the proposed
method. Stream processing is not versatile for accelerating any kinds of signal
processing: It is only feasible for computations that utilize simple data 
epeatedly and can compute each sub-process with almost the same calculation
cost. We modify the algorithm so as to make it plausible for stream processing.

The rest of the paper is organized as follows: Section \ref{sec:framework}
describes the framework of the proposed method. Section \ref{sec:saliency}
presents the procedure how to estimate human visual attention based on visual
saliency. Section \ref{sec:graphcuts} explains a technique for supervised image
segmentation based on graph cuts as a basis of our proposed method. Sections
\ref{sec:prior} to \ref{sec:implement} present our main contributions of this
paper, namely the method for providing saliency-based priors, the method for
updating the priors according to the previous segmentation result, and their
implementation based on the idea of stream processing. Section \ref{sec:eval}
discusses some quantitative evaluations. Finally, Section \ref{sec:conclude}
summarizes the paper and discusses future work.

\section{Framework}
\label{sec:framework}

\begin{figure}[t]
  \begin{center}
    \includegraphics[clip,width=0.7\hsize]{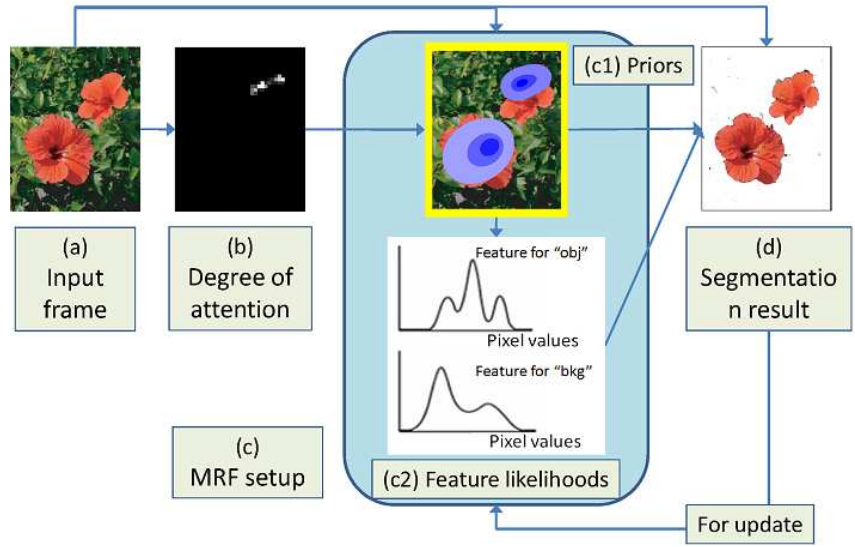}
    \caption{Framework of the proposed method}
    \label{fig:framework}
  \end{center}
\end{figure}

This section describes the framework of the proposed method for extracting
salient regions from videos. Figure \ref{fig:framework} depicts the framework.

First, the visual attention density is calculated from each frame of an input video via
a saliency-based human visual attention model. Although any kind of attention model can
be employed, we utilize the model proposed by Pang et al.
\cite{stochasticSaliency:derek,stochasticSaliencyMRF:derek} to compute the human visual
attention density. Section \ref{sec:saliency} describes how to estimate human visual
attention with the proposed method.

Next, a Markov random field (MRF) model for segmentation is prepared, where each
hidden state corresponds to the label of a position representing an ``object'' or
``background'', and an observation is a frame of the input frame. The density calculated
in the previous step can be utilized for estimating the priors of objects/backgrounds and
the feature likelihoods of the MRF. When calculating priors and likelihoods, the regions
extracted from the previous frames are also available. Sections \ref{sec:prior} and
\ref{sec:update} focus
particularly on how to determine and update priors and feature likelihoods based on the
density of visual attention and previous segmentation results.

Once the MRF is constructed, salient regions can be obtained as the MAP solution of the
MRF. When estimating the MAP solution, graph cuts based methods \cite{graphCuts:boykov2}
can be employed. Section \ref{sec:graphcuts} presents the Interactive Graph Cuts method
for image segmentation.

\section{Estimation of human visual attention}
\label{sec:saliency}

\clearpage
\begin{figure}[t]
  \begin{center}
    \includegraphics[clip,width=0.7\hsize]{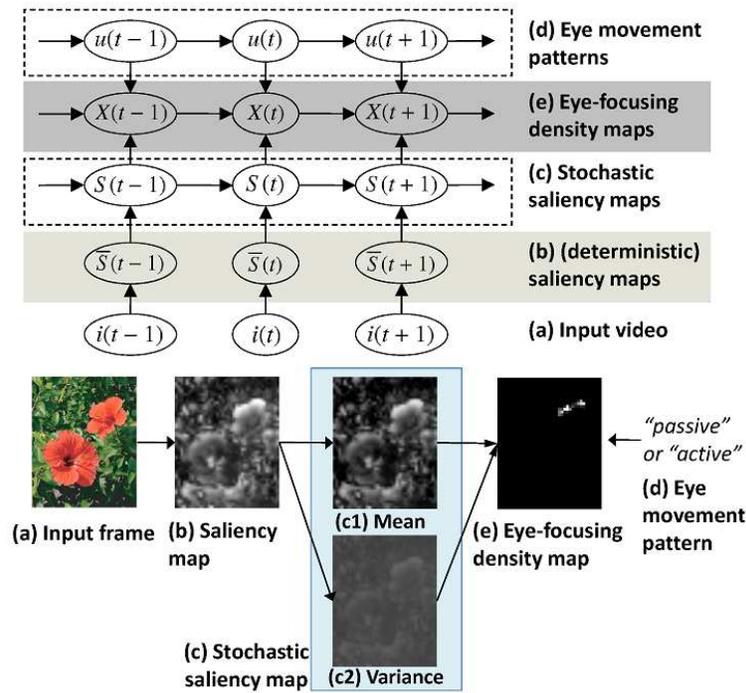}
    \caption{%
      Eye-focusing density estimation through a stochastic model of human visual attention
    }
    \label{fig:EFDM}
  \end{center}
\end{figure}

Figure \ref{fig:EFDM} shows the framework for estimating human visual
attention. We used the stochastic model of human visual attention proposed by Pang
et al. \cite{stochasticSaliency:derek,stochasticSaliencyMRF:derek}.

First, a saliency map is calculated from each frame of the input video with the method
proposed by Itti et al. \cite{Saliency:itti}. Our implementation utilized intensity,
color opponents, orientation and motion information as fundamental features.

Then, a stochastic representation of the saliency map is computed through a Kalman
filter, where the saliency map is utilized as the observation of the filter. We call the
stochastic representation of the saliency map as a stochastic saliency map. Each pixel
of the stochastic saliency map is expressed by a Gaussian density. 

The density of human visual attention can be directly calculated from the stochastic
saliency map by introducing the principle of the signal detection theory
\cite{SignalDetectionTheory:Eckstein}, namely, the position at which stochastic saliency
takes its maximum value is the eye focusing position. Since each pixel of the stochastic
saliency map is expressed by a Gaussian, we can calculate the visual attention density
for each pixel such that the saliency value has its maximum value at that pixel.

The model also incorporates another property, namely that eye movements may be affected
by a cognitive state. The cognitive state is represented as an eye movement pattern in
this model. Two typical eye movement patterns, passive and active, are found when a
person is watching a video. By introducing the eye movement patterns, eye movements can
be modeled with a hidden Markov model.

Finally, by integrating the density related to the bottom-up part (namely the stochastic
saliency map) and the top-down part (namely the eye movement pattern), we can obtain the
final density of visual attention, which is called the eye focusing density map (EFDM).

Although the above procedure well simulated the human visual system, it requires high
computational costs (about 1 second per frame with a standard workstation). When
considering this model as a pre-selection mechanism for subsequent
processing (i.e. video segmentation), computational cost should be of crucial significance
in terms of practical use. We have developed an algorithm plausible for stream processing
\cite{stochasticSaliencyGPU}, which incorporates a particle filter
\cite{beyondTheKalmanFilter} with Markov chain Monte-Carlo sampling
\cite{IntroductionMCMC:andrieu} into the basic model \cite{stochasticSaliency:derek}.
Details can be seen in \cite{stochasticSaliencyGPU,stochasticSaliency:derek_paper}.

\section{Segmentation with graph cuts}
\label{sec:graphcuts}

This section describes the supervised image segmentation technique based on graph cuts
proposed by Boykov et al. \cite{graphCuts:boykov2}.

We start by describing MRFs for image segmentation. Consider a set of random variables
$\A=\{A_{\x}\}_{\x\in I}$ defined on a set $I$ of coordinates. Each random variable
$A_{\x}$ takes a value $a_{\x}\in\{0,1\}$ corresponding to a
background ($0$) and an object ($1$). Its inference can be
formulated as an energy minimization problem where the energy corresponding to the
configuration $\A$ is the negative log likelihood of the posterior density of the
MRF, $E(\A|\D)=-\log p(\A|\D)$, where $\D$ represents the input image. The energy
function consists of likelihood and prior terms defined as follows:
\begin{eqnarray}
  \lefteqn{E(\A|\D)
   =  \sum_{\x\in I}\left\{\psi_1(\D|A_{\x})+\xi_1(A_{\x})\right.}\nonumber\\
  &+& \sum_{\y\in N_{\x}}
      \left(\psi_2(\D|A_{\x},A_{\y})+\xi_2(A_{\x},A_{\y})\right)\},
      \label{eq:graphcuts:energy_total}
\end{eqnarray}
where $N_{\x}$ is a neighboring system for the position $\x$, $\psi_i(\D|\cdot)$
$(i=1,2)$ is a likelihood term and $\xi_i(\cdot)$ is a prior term. The first
likelihood term $\psi_1(\D|A_{\x})$ imposes individual penalties for assigning
label $a\in\{0,1\}$ to pixel $\x$, and it is given by $\psi_1(\D|A_{\x})=%
-\log p(C_{\x}|A_{\x})$, where $C_{\x}$ is the RGB value at the position $\x$.
The likelihood $p(C_{\x}|A_{\x})$ of the RGB values can be modeled as a Gaussian
mixture model (GMM), and estimated with a standard EM algorithm, where the
number of Gaussians is given in advance as $M$. The first prior term
$\xi_1(A_{\x})$ represents how the position is likely to a object, and can be
determined by label manually given from users as
\begin{eqnarray*}
  \xi_1(A_{\x}) &=& -\log p(A_{\x}),\\
  p(A_{\x}=1) &=& \left\{
  \begin{array}{ll}
    1   & \x\mbox{ has a manual label }1,\\
    \epsilon\approx 0 & \x\mbox{ has a manual label }0,\\
    0.5 & \mbox{No label provided at }\x
  \end{array}\right.\\
  p(A_{\x}=0) &=& 1-p(A_{\x}=1).
\end{eqnarray*}
The second prior term $\xi_2(A_{\x},A_{\y})$ takes the form of a generalized
Potts model as $\xi_2(A_{\x},A_{\y})=constant$ only if $A_{\x}\neq A_{\y}$. The
second likelihood term $\psi_2(D|A_{\x},A_{\y})$ reduces the cost for two
labels, which differs in proportion to the difference between the intensity
values of their corresponding positions.
\begin{eqnarray*}
  \psi_2(D|A_{\x},A_{\y}) &\propto&
  -\exp\left\{ -\frac{(I_{\x}-I_{\y})^2}{2\sigma^2}\right\}
  \cdot\frac{1}{\|\x-\y\|}\\
  & & \mbox{if }A_{\x}\neq A_{\y}
\end{eqnarray*}
where $I_{\x}$ denotes the intensity at the pixel $\x$.

\begin{figure}[t]
  \begin{center}
    \includegraphics[clip,width=0.7\hsize]{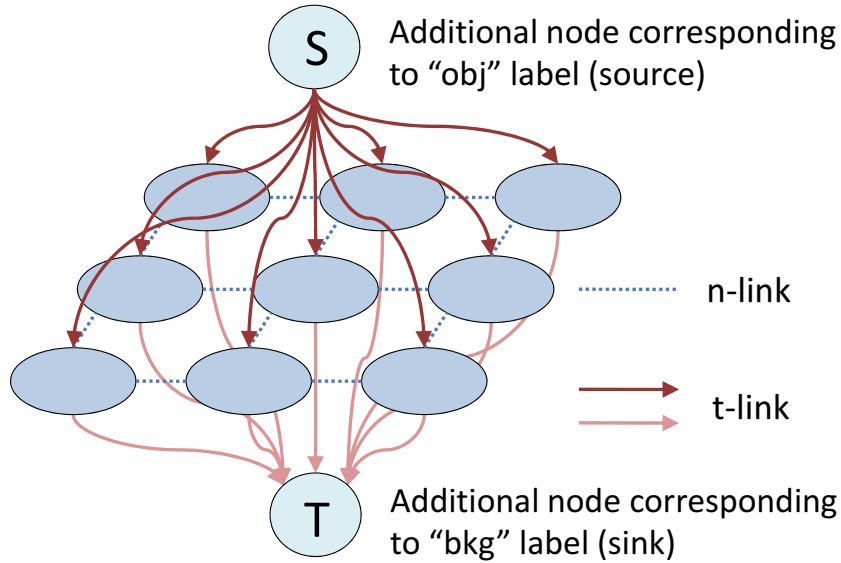}
    \caption{Graph for image segmentation}
    \label{fig:graph}
  \end{center}
\end{figure}

The MRF configuration $\wh{\a}$ with the least energy corresponds to the MAP solution
of the MRF. The energy minimization can be performed by finding the minimum cut on
an equivalent graph of the MRF as shown in Figure \ref{fig:graph}. Each random variable
$A_{\x}$ of the MRF is represented by a vertex $v_{\x}$ in this graph. A directed edge from
each vertex $v_{\x}$ is connected to another vertex in its neighborhood $N_{\x}$%
\footnote{Eventually, each pair of neighboring vertices has a pair of mutually connected
directed edges. Thus, we represent the pair of directed edges as an undirected edge in
Figure \ref{fig:graph}.}. These edges are called as neighborhood links
(n-links). The cost $c(v_{\x},v_{\y})$ associated with the n-link $(\x,\y)$ connecting
from $v_{\x}$ to $v_{\y}$ is given by the sum of the second prior and likelihood
terms as
\begin{eqnarray}
c(v_{\mbox{\boldmath $x$}}, v_{\mbox{\boldmath $y$}}) &=& 
\psi_2(D|A_{\mbox{\boldmath $x$}},
A_{\mbox{\boldmath $y$}}) + \xi_2(A_{\mbox{\boldmath $x$}},A_{\mbox{\boldmath $y$}}).
\end{eqnarray}
Also, the graph has two special vertices the source $s$ and the sink $t$ each of which
corresponds to the label $0$ and $1$. Directed edges called terminal links
(t-links) are connected from the source to all the other vertices except the sink and
from all the vertices except the source to the sink. The costs $c(s,\vv_{\x})$ and
$c(t,\vv_{\x})$ of t-links are given by the sum of the first prior and likelihood terms as
\begin{eqnarray}
  c(s, v_{\mbox{\boldmath $x$}}) &=& \psi_1(D|A_{\mbox{\boldmath $x$}}=0) 
					+ \xi_1(A_{\mbox{\boldmath $x$}}=0),\\
  c(v_{\mbox{\boldmath $x$}},t) &=& \psi_1(D|A_{\mbox{\boldmath $x$}}=1)
					+ \xi_1(A_{\mbox{\boldmath $x$}}=1).
\end{eqnarray}
The minimum cut of the graph separating the source and the sink provides the MAP
configuration $\wh{\a}$ of the corresponding MRF.

\section{Saliency-based priors}
\label{sec:prior}

\begin{figure}[t]
  \begin{center}
    \includegraphics[clip,width=0.7\hsize]{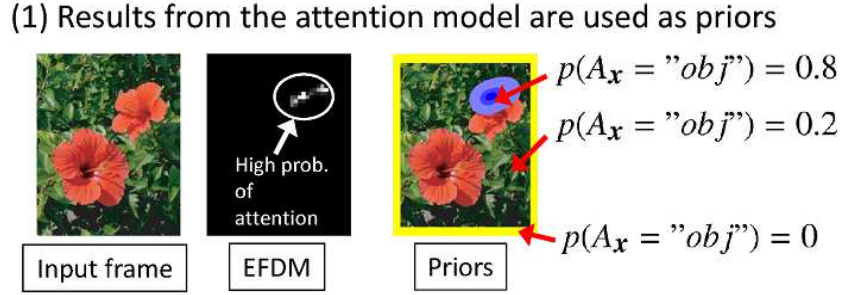}
    \caption{Saliency-based priors}
    \label{fig:label1}
  \end{center}
\end{figure}

As the first contribution of this paper, we provide a way to calculate the first prior
term of the energy function shown in Equation (\ref{eq:graphcuts:energy_total}) without
any manually provided labels. We utilize the density of visual attention calculated by the
procedure shown in Section \ref{sec:saliency}. Figure \ref{fig:label1} shows a sketch for
calculating the prior. 

The prior density $p(A_{\x}=1)$ is obtained from the EFDM (cf. Section \ref{sec:saliency}).
We represent the EFDM with a Gaussian mixture model (GMM), and
estimate the model parameter with the EM algorithm. The estimated GMM density
represents the prior density $p(A_{\x}=1)$. Exceptionally, the prior on the edge
of each frame is assumed to be $p(A_{\x}=1)\approx 0$ since some of the background
regions are expected to be at the frame edge.

The likelihood density $p(C_{\x}|A_{\x})$ can be obtained in the same way as the
Interactive Graph Cuts \cite{graphCuts:boykov2}. Although in the Interactive Graph Cuts,
samples are selected from the manually-labeled pixels for estimating the likelihood density
$p(C_{\x}|A_{\x})$, our proposed method utilizes all the pixels, where samples are weighted
by the prior density $p(A_{\x}=1)$.

\section{Prior update}
\label{sec:update}

\begin{figure}[t]
  \begin{center}
    \includegraphics[clip,width=0.7\hsize]{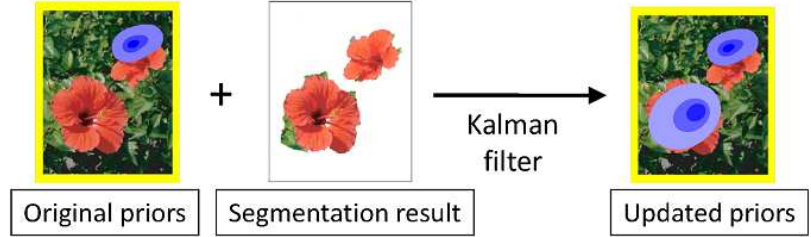}
    \caption{Prior update}
    \label{fig:label2}
  \end{center}
\end{figure}

The second contribution provided by our method is that it offers a way to update the prior
and likelihood terms according to the segmentation results derived from the previous
frames and the density of visual attention calculated from the current frame.
Figure \ref{fig:label2} shows a sketch for prior update.  Here, we
introduce a notation $\A_t=\{A_{\x,t}\}_{\x\in I}$ $(t=0,1,\cdots)$ for representing the
MRF configuration at time $t$.

To update the prior density $p(A_{\x,t})$ at time $t$, we introduce an idea of Kalman
filter \cite{beyondTheKalmanFilter}, where the prior density derived solely from the EFDM
at time $t$ (from now on, we denote it as $q(A_{\x,t})$) is considered to be the
observation at time $t$. We assume the following two relationships:
\begin{eqnarray*}
  p(A_{\x,t}=1) &=& f(\wh{\A}_{t-1},\x) + N_1,\\
  q(A_{\x,t}=1) &=& p(A_{\x,t}=1) + N_2,
\end{eqnarray*}
where $\wh{\A}_t$ is the estimated MRF configuration at time $t$, $N_i$ $(i=1,2)$ is a
Gaussian random variable with mean $0$ and variance $\sigma^2_i$, and  $f(\A,\x)$
represents a pixel
value at $\x$ of a gray-scaled image obtained from an MRF configuration $\A$
with some image processing e.g. Gaussian smoothing \cite{multiScaleGraphCuts:nagahashi} or
distance transform \cite{conf_icme_FukudaTA08}. These equations imply that the
prior density at the current frame depends on both the EFDM at the
current frame and the segmentation result of the previous frame.

The estimate $\wh{p}(A_{\x,t}=1)$ of the prior density at time $t$ can be derived as
\begin{eqnarray}
\widehat{p}(A_{\mbox{\boldmath $x$},t} = 1) &=& \frac{\sigma ^2_1}
	{\sigma ^2_1 + \sigma ^2_2 + \sigma ^2_{\xi _1}(t-1)} f(\mbox{\boldmath $\widehat{A}$}_{t-1},\mbox{\boldmath $x$}) \nonumber\\
	&+& \frac{\sigma ^2_2 + \sigma ^2_{\xi _1}(t-1)}
	{\sigma ^2_1 + \sigma ^2_2 + \sigma ^2_{\xi _1}(t-1)}q(A_{\mbox{\boldmath $x$},t}=1),\nonumber\\
\sigma ^2_{\xi _1}(t) &=& \frac{\sigma ^2_1\cdot (\sigma ^2_2 + \sigma ^2_{\xi _1}(t-1))}
	{\sigma ^2_1 + \sigma ^2_2 + \sigma ^2_{\xi _1}(t-1)}.\nonumber
\end{eqnarray}
The estimate $\wh{p}(A_{\x,t}=1)$ derived from the above procedure is used as a
new prior density.

\section{GPU implementation}
\label{sec:implement}

\subsection{Stream processing}
\label{sec:implement:intro}

In recent years, there has been strong interest from researchers and developers in
exploiting the power of commodity hardware including multiple processor cores for
parallel computing. This is because 1) multi-core CPUs and stream processors such as
graphics processing units (GPUs) and Cell processors \cite{CELL} are currently the most
powerful and economical computational hardware available, 2) the rise of SDKs and APIs
such as NVIDIA CUDA \cite{CUDA}, AMD ATiStream \cite{ATI_stream}, OpenCL \cite{OpenCL} and
Microsoft DirectCompute \cite{DirectX} makes it easy to implement
desired algorithms for execution on multi-core hardware. This programming paradigm is
widely known as \textit{stream processing}. However, stream processing is
not versatile for accelerating any kinds of signal processing: Stream processing is only
feasible for computations that utilize simple data repeatedly and can compute each
sub-process with almost the same calculation cost. When we make extensive use of stream
processing, we have to modify the algorithm to fit the above property.

We are focusing on GPUs as prospective hardware for stream processing, due to its powerful
performance and availability. Previously, we needed to master shader programming languages
such as HLSL \cite{DirectX} and GLSL \cite{OpenCL} as well as to understand graphics
pipelines for the extensive use of GPUs. NVIDIA CUDA \cite{CUDA} makes it easy to implement
a wide variety of (numerical, now always graphics-related) algorithms without any special
knowledge and artifices. Its interface is quite similar to
C, and its function can be called in standard C/C++ platforms.

\begin{figure}[t]
  \begin{center}
    \includegraphics[width=0.7\hsize]{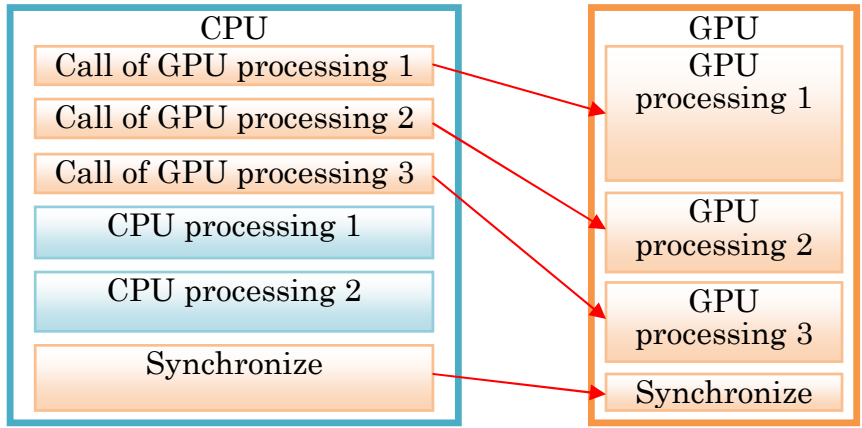}
    \caption{Asynchronous execution of CPU and GPU processing}
    \label{fig:cuda_execute_timing}
  \end{center}
\end{figure}

\begin{figure}[t]
  \begin{center}
    \includegraphics[width=0.7\hsize]{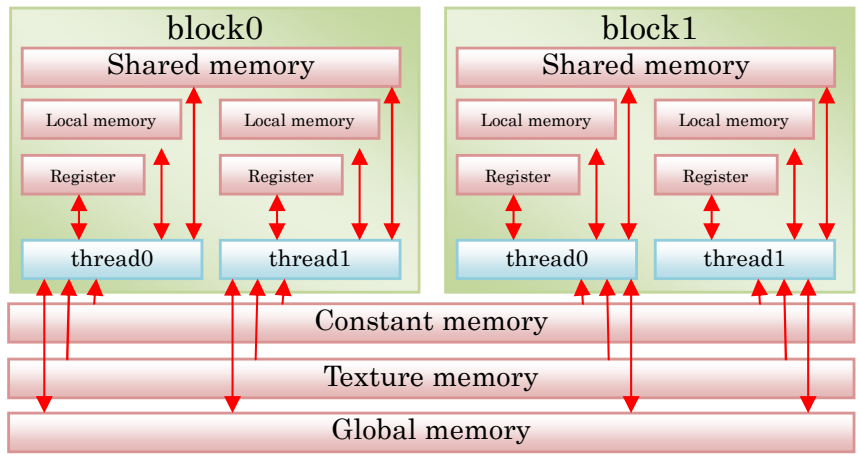}
    \caption{Memory model on CUDA}
    \label{fig:cuda_memory_model}
  \end{center}
\end{figure}

\begin{table}
  \begin{center}
    \caption{Characteristics of physical memories on GPU}
    \label{tab:cuda_memory_feature}
    \begin{tabular}{c|c|c|c|c|c}\hline
      Type & Read & Write & Data & Cache & Synchro- \\
      & & & transfer & & nization \\ \hline\hline
      Global & OK & OK & OK & NG & Grids \\ \hline
      Texture & OK & NG & OK & OK & --- \\ \hline
      Constant & OK & NG & OK & OK & --- \\ \hline
      Shared & OK & OK & NG & --- & Blocks \\ \hline
      Local & OK & OK & NG & --- & --- \\ \hline
      Register & OK & OK & NG & --- & --- \\ \hline
    \end{tabular}
  \end{center}
\end{table}

Although CUDA enables us to implement various kinds of algorithms easier than before, we
should still take care of its programming model and memory model for the extensive use. As
shown in Figure \ref{fig:cuda_execute_timing}, once GPU processing is called from a CPU
functions, it is queued in line by a graphics driver and executed sequentially and
asynchronously. This implies that excess computational resources of CPU can be assigned to
other computations such as data transfer between CPU and GPU. 
Also, as shown in Figure \ref{fig:cuda_memory_model} and Table
\ref{tab:cuda_memory_feature}, CUDA can handle 6 different types of memories: Global,
texture, constant, shared, local and register. Moreover, data transfer between CPU and GPU
often becomes the bottle neck for the acceleration. From the above discussion, we have to
carefully consider the order and timing of function calls for GPU, and select the type of
memories according to the usage.

\subsection{Visual attention estimation}
\label{sec:implement:saiency}

Almost all the parts for estimating visual attention have been already implemented on GPU,
however, saliency map calculation \cite{Saliency:itti} still remains as a CPU processing.
This section details how to implement saliency map calculation on GPU.

Saliency map calculation consists of 1) fundamental feature extraction such as intensity,
color opponents, edge orientation and optical flow, 2) Gaussian pyramid construction, 3)
a special normalization function utilizing the global and local minimum of pixel values,
and 4) weighted addition of images. These computation can be roughly classified into the
following 3 types: pixel-wise computation, filter convolution, and local extrema detection.
In the following, we detail each procedure of.

\begin{figure}[t]
  \begin{center}
\begin{lstlisting}
texture<float, 1, cudaReadModeElementType> filter;
texture<float, 1, cudaReadModeElementType> src1;
__device__ float Filter2DCore(texture<float, 1, cudaReadModeElementType> fsource, int x, int y, int height, int width, int filterSizeX, int filterSizeY) {
	float sum = 0;
	x -= filterSizeX/2;
	y -= filterSizeY/2;
	for(int fy = 0; fy < filterSizeY; fy++) {
		int by = y + fy;
		if(by > 0 && by < height) {
			by *= width;
			for(int fx = 0; fx < filterSizeX; fx++) {
				int bx = x + fx;
				if(bx > 0 && bx < width) {
					sum += tex1Dfetch(filter, fy*filterSizeX + fx)*tex1Dfetch(fsource, by + bx);
				}
			}
		}
	}
	return sum;
}
__global__ void Filter2DKernel(int height, int width, int fheight, int fwidth, float* result) {
	int px = blockDim.x*blockIdx.x + threadIdx.x;
	if(px < height*width) {
		int x = px%width;
		int y = px/width;
		result[px] = Filter2DCore(src1, x, y, height, width, fwidth, fheight);
	}
}
\end{lstlisting}
    \caption{CUDA implementation for filtering}
    \label{fig:filter}
  \end{center}
\end{figure}

A pixel value $F(\x)$ of the filtered image at the position $\x$ can be derived by
convoluting the original image $\D$ with a filter kernel $F_k(\x)$ with size $n\times m$
as follows:
\begin{eqnarray*}
  F(x,y) &=& \sum_{i=0}^{n}\sum_{j=0}^{m}F_k(i,j)\\
         & & *P\left(x+i-\frac{n}{2}, y+j-\frac{m}{2}\right).
\end{eqnarray*}
The image $P(x,y)$ and filter kernel $F_k(i,j)$ are transferred to and placed on the texture
memory, and the filter output $F(x,y)$ is set on the global memory. This allocation would
enhance the performance of memory access since the filter kernel $F_k(i,j)$ are utilized by
every kernel and every pixel $P(x,y)$ of the image is accessed by several threads.
A pseudo code for filter convolution is shown in Figure \ref{fig:filter}.

Gaussian pyramids can be efficiently constructed by setting the image on the texture
memory.

\begin{figure}[!t]
  \begin{center}
\begin{lstlisting}
texture<float, 1, cudaReadModeElementType> minsrc;
texture<float, 1, cudaReadModeElementType> maxsrc;
__global__ void SMRangeNormalizeKernel1(int length, float* localmin, float* localmax) {
	int px = blockDim.x*blockIdx.x + threadIdx.x;
	
	__shared__ float mini[32], maxi[32];
	if(px < length) {
		mini[threadIdx.x] = tex1Dfetch(minsrc, px);
		maxi[threadIdx.x] = tex1Dfetch(maxsrc, px);
	} else {
		mini[threadIdx.x] = FLT_MAX;
		maxi[threadIdx.x] = FLT_MIN;
	}
	__syncthreads();
	if(threadIdx.x == 0) {
		for(int i = 1; i < blockDim.x; i++) {
			mini[0] = min(mini[0], mini[i]);
			maxi[0] = max(maxi[0], maxi[i]);
		}
		localmin[blockIdx.x] = mini[0];
		localmax[blockIdx.x] = maxi[0];
	}
}
\end{lstlisting}
    \caption{Minimum/maximum search}
    \label{fig:search_min_max}
  \end{center}
\end{figure}

Figure \ref{fig:search_min_max} shows a pseudo code for searching the global and local
extrema of pixel values in the image, where a buffer for the minima (resp. maxima) is
denoted as {\emph minsrc} (resp. {\emph maxsrc}). Every pixel value in a block is first
obtained and stored in the shared memory. Then, all the thread in the block are
synchronized by calling the function {\emph \_\_syncthreads}, and some specific thread
(e.g. thread 0) computes the maximum and minimum in the block. As a result, a smaller image
having the same number of pixels as the number of blocks and pixel values of the minimum
(resp. minimum) of every blocks is generated. This procedure is repeatedly executed until
the number of pixels converges.

\subsection{Segmentation}
\label{sec:implement:segmentation}

For the segmentation procedure, we newly implement the algorithm for deriving priors,
feature likelihoods and minimum cuts on GPU. 

Priors can be calculated in almost the same way as described in the previous section, since
the procedure is composed of Gaussian filtering and pixel-wise Kalman filter.

For the derivation of t-link costs, we first estimate GMM model parameters of RGB values
(see Section \ref{sec:graphcuts}) with EM algorithm, which has been already implemented
and distributed by Harp \cite{GMM_CUDA:harp}. We utilized k-means algorithm implemented on
CPU for the initialization of the EM algorithm. T-link costs are the negative log
likelihood of image features, which can be implemented on GPU by a combination of Gaussian
filtering. Only the normalization term of each Gaussian density is calculated on CPU.

For the graph cuts, we can find several implementations with CUDA. We utilized CUDA Cuts
\cite{CUDAcuts} developed by Vineet et al.

\clearpage
\section{Experiments}
\label{sec:eval}

\subsection{Conditions}
\label{sec:eval:condition}

\begin{figure}[t]
  \begin{center}
    \includegraphics[clip,width=0.7\hsize]{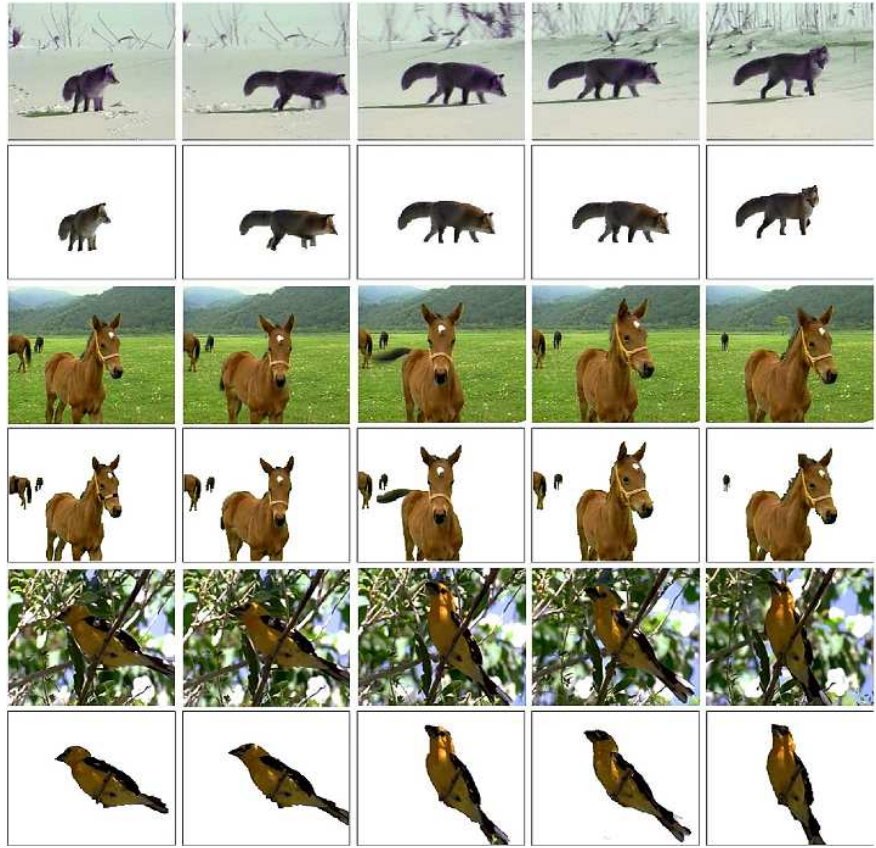}
    \caption{Samples of input videos and corresponding ground truth made by hands}
    \label{fig:correctdata}
  \end{center}
\end{figure}

To verify the effectiveness of the proposed method, we conducted video segmentation for
10 video clips of length 5-10 seconds and 12 fps. For each video, we have made ground-%
truth segmented video frames by hands. Some examples can be seen in Figure
\ref{fig:correctdata}. As a measure for quantitative evaluation, we adopted error rate, 
precision, recall and F-value defined as follows:
\begin{eqnarray*}
  {\rm Error} &=& \frac{FP + FN}{TP + TN + FP + FN},\\
  {\rm Recall} &=& \frac{TP}{TP + FN},\\
  {\rm Precision} &=& \frac{TP}{TP + FP},\\
  {\rm F-value} &=& \frac{2\times {\rm Recall}\times {\rm Precision}}
                         {{\rm Recall} + {\rm Precision}},
\end{eqnarray*}
where TP, TN, FP and FN respectively represents the number of true positives, true
negatives, false positives and false negatives. We compared our new method with the
following methods
\begin{enumerate}
\item Manual: Manually-provided labels were available for the segmentation only in the
first frame. For the other frames, priors and feature likelihoods were estimated from the
previous segmentation result. This strategy is quite similar to the semi-automatic method
developed by Kohli and Torr \cite{graphCuts:kohliPaper}.
\item Non-update: Only saliency-based priors were available and any previous segmentation
results cannot be utilized for the segmentation. This strategy simulates the fully
automatic method developed for still images by Fu et al. \cite{saliencyCuts:fu}.
\item Update: Our proposed method
\end{enumerate}
\begin{table}
  \begin{center}
    \caption{Platform used in the evaluation}
    \label{tab:platform}
    \begin{tabular}{c|c}\hline
      CPU & Intel Core2Quad Q9550 \\ \hline
      Memory & 4GB \\ \hline
      GPU & NVIDIA Geforce 9800GT \\ \hline
      Graphics memory & 512MB \\ \hline
      OS & Windows XP Professional \\ \hline
      Software & NVIDIA CUDA 2.1 \\
      & OpenCV 1.1pre \\ \hline
    \end{tabular}
  \end{center}
\end{table}
We experimentally determined parameters in advance as follows: 
$\sigma _1 = 0.03, \sigma _2 = 0.035, M = 3$.
The platform used in the evaluation is shown in Table \ref{tab:platform}.

\subsection{Segmentation accuracy}
\label{sec:eval:accurate}

\begin{figure}[t]
  \begin{center}
    \includegraphics[clip,width=0.7\hsize]{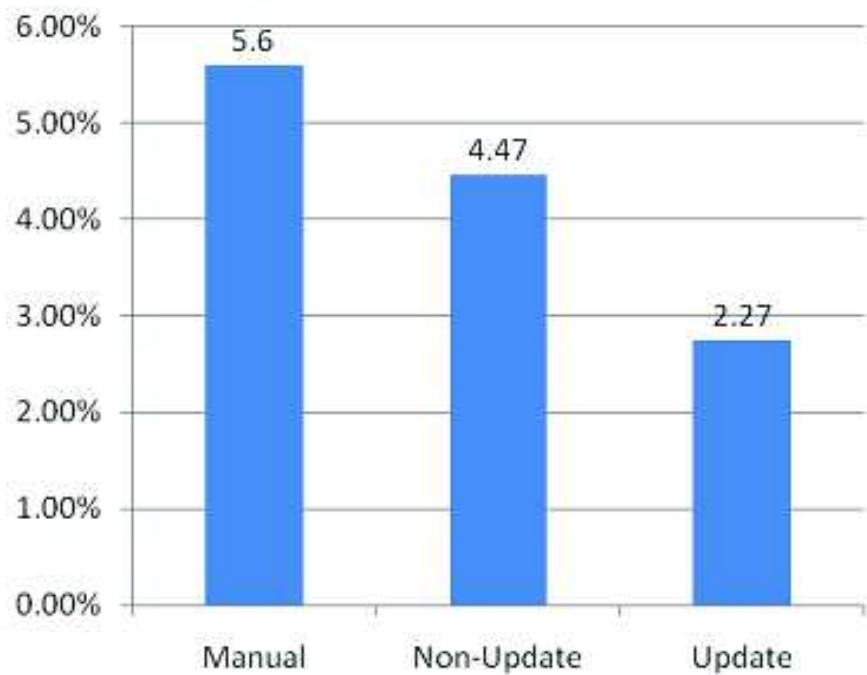}
    \caption{Evaluation result measured by the error rate}\label{fig:error}
  \end{center}
\end{figure}

\begin{figure}[t]
  \begin{center}
    \includegraphics[clip,width=0.7\hsize]{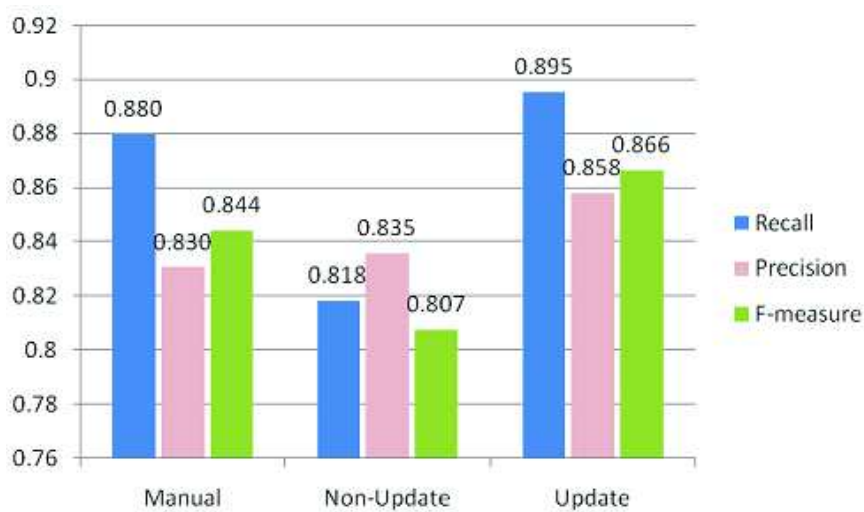}
    \caption{Evaluation result measured by recall, precision and F-measure}
    \label{fig:segmentation_result}
  \end{center}
\end{figure}

Figure \ref{fig:error} shows the segmentation accuracy measured by the error rate, and
Figure \ref{fig:segmentation_result} shows the accuracy measured by precision, recall and
F-measure. These results indicates that our proposed method outperformed the other methods
under all the conditions.

\begin{figure}[t]
  \begin{center}
    \includegraphics[clip,width=0.7\hsize]{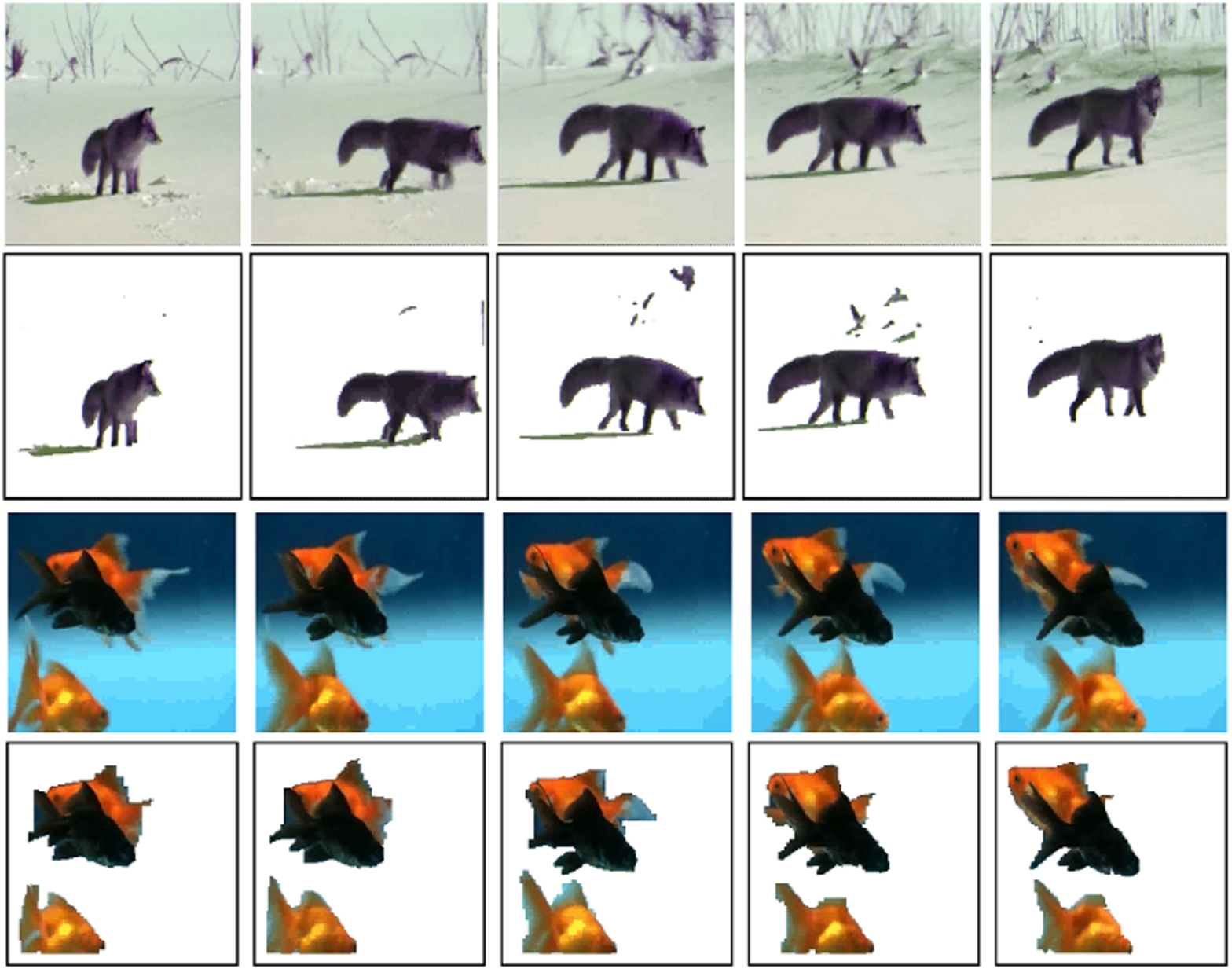}
    \caption{Segmentation examples emitted from the proposed method}
    \label{fig:seg_results}
  \end{center}
\end{figure}

Figure \ref{fig:seg_results} shows some examples emitted from our proposed method. By 
comparing it with Figure \ref{fig:correctdata}, we can see that our proposed method worked
well from the qualitative aspect.

\begin{figure*}[t]
  \begin{center}
    \includegraphics[clip,width=\hsize]{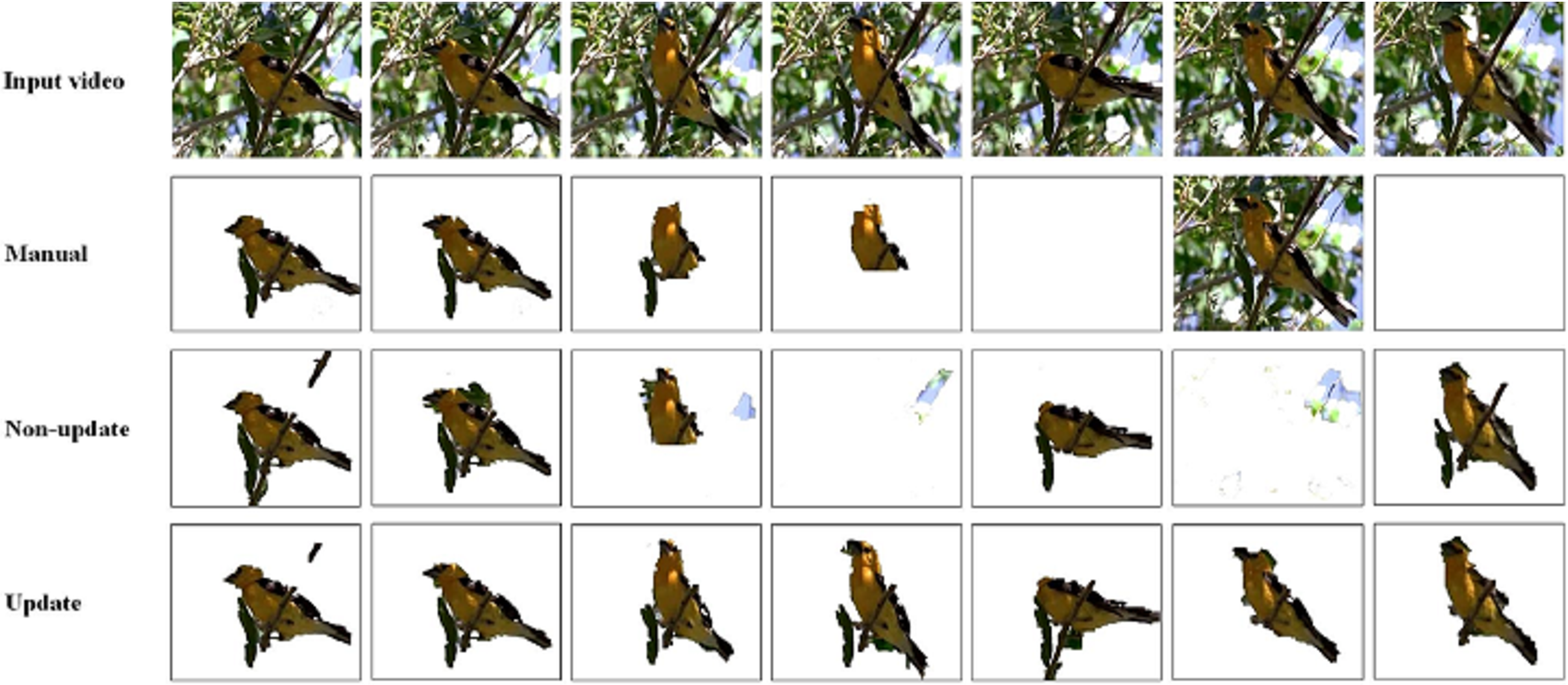}
    \caption{Comparing segmentation results emitted from all the 3 methods}
    \label{fig:compare}
  \end{center}
\end{figure*}

Figure \ref{fig:compare} shows an example of segmentation results emitted from all the
methods used in this evaluation. The method ``manual'' could not recover from incorrect
segmentation once the target (in this case a bird) lost, since this method only utilized
the previous segmentation results as a cue for detecting the target. This indicates the
advantages of saliency-based priors. The segmentation results emitted from the method
``non-update'' sometimes became unstable due to some noises or fluctuations included in
saliency information. This implies that temporal smoothness by utilizing the previous
segmentation result is also significant for stable segmentation.

\begin{table}
  \begin{center}
    \caption{Comparing error rates [\%]}
    \label{tab:error}
    \begin{tabular}{|l|c|c|c|} \hline
      & Boykov \cite{graphCuts:boykov1} & Nagahashi \cite{multiScaleGraphCuts:nagahashi} & Proposed \\ \hline
      Error & 3.75 & 1.61 & 2.74\\ \hline
    \end{tabular}
  \end{center}
\end{table}

\begin{table}
  \begin{center}
    \caption{Comparing recall, precision and F-measure}
    \label{tab:seido}
    \begin{tabular}{|l|c|c|c|} \hline
      & Boykov \cite{graphCuts:boykov2} & Nagahashi \cite{spatioTemporalGraphCuts:nagahashi} & Proposed \\ \hline
      Recall 	  & 0.88 & 0.91 & 0.895 \\\hline
      Precision & 0.96 & 0.88 & 0.858 \\\hline
      F-value & 0.92 & 0.89 & 0.866 \\\hline
    \end{tabular}
  \end{center}
\end{table}

We show some reference information as to the segmentation accuracy. Table \ref{tab:error}
presents error rates published in the papers of Boykov et al. \cite{graphCuts:boykov1} and
Nagahashi et al. \cite{multiScaleGraphCuts:nagahashi}, both of which are specialized for
still image segmentation with manually provided labels. Table \ref{tab:seido} shows
precision, recall and F-measure published in the paper of Boykov et al.
\cite{graphCuts:boykov2} and Nagahashi et al. \cite{spatioTemporalGraphCuts:nagahashi},
both of which are specialized for video segmentation with manually provided labels. 
This table indicates that our proposed method marked high segmentation accuracy comparable
with the previously proposed semi-automatic segmentation methods. Note that videos used for
the evaluation differs from each other.

\subsection{Execution time}
\label{sec:eval:time}

\begin{figure}[t]
  \begin{center}
    \includegraphics[width=0.7\hsize]{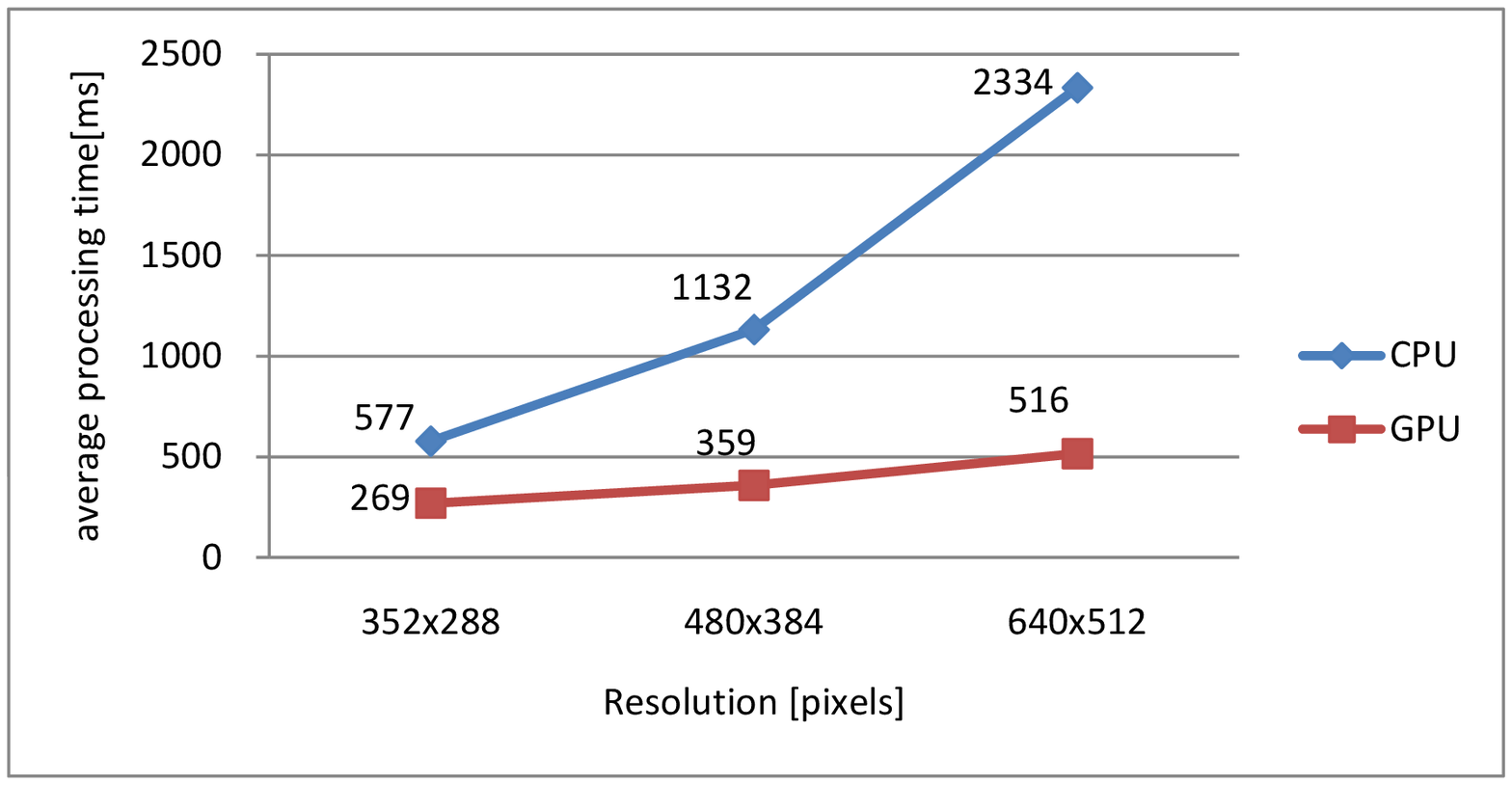}
    \caption{Average execution time per frame}
    \label{fig:avg_proc_time_resol}
  \end{center}
\end{figure}

\begin{table}
  \centering
    \caption{Detailed execution time per frame [ms]}
    \label{tab:avg_proc_time}
    \begin{tabular}{c|c|c|c|c|c|c}\hline
      & & VA & Priors & t-link & Graph & Misc \\
      & &    &        &        & cuts  & \\ \hline\hline
      352 & CPU & 32.9 & 148.1 & 218.6 & 97.0 & 71.0 \\\cline{2-7}
      $\times$288 & GPU & 22.2 & 1.9 & 109.6 & 69.0 & 65.6 \\\hline
      480 & CPU & 58.8 & 372.8 & 350.8 & 246.5 & 86.4 \\\cline{2-7}
      $\times$384 & GPU & 30.4 & 3.5 & 120.8 & 27.7 & 74.6 \\\hline
      640 & CPU & 109.8 & 814.5 & 602.6 & 664.5 & 112.7 \\\cline{2-7}
      $\times$512 & GPU & 45.2 & 6.2 & 142.6 & 232.3 & 87.1 \\\hline
    \end{tabular}
\end{table}

Figure \ref{fig:avg_proc_time_resol} shows the average execution time per frame for the
cases of CPU and GPU implementations, and Table \ref{tab:avg_proc_time} shows the detailed
execution time per frame for each step, where ``misc'' includes the time for capturing
video frames, memory allocation and release. These results indicate that GPU implementation
greatly improve the execution time, e.g. 132 times in deriving saliency-based priors and
4.5 times in total than the CPU implementation. These
results also indicates that as the video resolution increased the execution time per pixel
decreased in the GPU implementation, while the opposite in the CPU implementation.

\begin{table}
  \caption{Comparison of execution time with previous methods}
  \label{tab:speed}
  \begin{center}
    \begin{tabular}{|c|c|c|c|c|} \hline
      \cite{graphCuts:boykov1} & \cite{multiScaleGraphCuts:nagahashi} &
      \cite{graphCuts:boykov2} & \cite{spatioTemporalGraphCuts:nagahashi} &
      Proposed\\
      300$\times $255 & 300$\times $255 & 360$\times $240 & 360$\times $240 &
      352$\times $288 \\ \hline
      0.94 & 37.59 & 329.6 & 181.6 & 0.294 \\ \hline
    \end{tabular}
  \end{center}
\end{table}

We show some reference information also as to the execution time. Table \ref{tab:speed}
presents the execution time published in the papers of Boykov et al.
\cite{graphCuts:boykov1,graphCuts:boykov2} and Nagahashi et al.
\cite{multiScaleGraphCuts:nagahashi,spatioTemporalGraphCuts:nagahashi}.
This table indicates that our proposed method finished all the procedures much faster than
others. Again note that videos used for the evaluation differs from each other.

\section{Conclusion}
\label{sec:conclude}

We have proposed a new and quite fast method for automatic video segmentation with the
help of 1) efficient optimization of Markov random fields with polynomial time of number
of pixels by introducing graph cuts, 2) automatic, computationally efficient but stable
derivation of segmentation priors using visual saliency and sequential update mechanism,
and 3) an implementation strategy in the principle of stream processing with graphics
processor units (GPUs). Experimental results indicated that our method extracted
appropriate regions from videos as precisely as and much faster than previous semi-%
automatic methods even though any supervisions have not been incorporated. Future work
includes development of more sophisticated segmentation methods utilizing such as top-down
information or text information.

\section*{Acknowledgment}
The authors also thank Dr. Naonori Ueda, Dr. Eisaku Maeda, Dr. Junji Yamato
and Dr. Kunio Kashino of NTT Communication Science Laboratories for their help.


\end{document}